# A Meaning-based Statistical English Math Word Problem Solver

**Chao-Chun Liang, Yu-Shiang Wong, Yi-Chung Lin and Keh-Yih Su**
Institute of Information Science, Academia Sinica, Taiwan
{ccliang, yushiangwtw, lyc, kysu}@iis.sinica.edu.tw

## Abstract

We introduce *MeSys*, a *meaning-based* approach, for solving English math word problems (MWPs) via *understanding and reasoning* in this paper. It first analyzes the text, transforms both body and question parts into their corresponding logic forms, and then performs inference on them. The associated context of each quantity is represented with proposed *role-tags* (e.g., *nsubj*, *verb*, etc.), which provides the flexibility for annotating an extracted *math quantity* with its associated context information (i.e., the *physical meaning* of this quantity). Statistical models are proposed to select the operator and operands. A noisy dataset is designed to assess if a solver solves MWPs mainly via understanding or mechanical pattern matching. Experimental results show that our approach outperforms existing systems on both benchmark datasets and the noisy dataset, which demonstrates that the proposed approach understands the meaning of each quantity in the text more.

## 1 Introduction

The *math word problem* (MWP) (see Figure 1) is frequently chosen to study natural language understanding and simulate human problem solving (Bakman, 2007; Hosseini et al., 2014; Liang et al., 2016) for the following reasons: (1) the answer to the MWP cannot be simply extracted by performing keyword/pattern matching. It thus shows the merit of understanding and inference.

| Math Word Problem |
|---|
| Mike takes 88 minutes to walk to school. If he rides a bicycle to school, it would save him 64 minutes. How much time did Mike save? |
| **Solution** |
| 88 – 64 = 22 |

Figure 1: An example of math word problem.

(2) An MWP usually possesses less complicated syntax and requires less amount of domain knowledge, so the researchers can focus on the task of understanding and reasoning. (3) The body part of MWP that provides the given information for solving the problem consists of only a few sentences. The understanding and reasoning procedures thus could be more efficiently checked. (4) The MWP solver has its own applications such as *Computer Math Tutor* (for students in primary school) and *Helper for Math in Daily Life* (for adults who are not good in solving mathematics related real problems).

According to the approaches used to identify entities, quantities, and to select operations and operands, previous MWP solvers can be classified into: (1) Rule-based approaches (Mukherjee and Garain, 2008[1]; Hosseini et al., 2014), which make all related decisions based on a set of rules; (2) Purely statistics-based approaches (Kushman et al., 2014; Roy et al., 2015; Zhou et al., 2015; Upadhyay et al., 2016), in which all related decisions are done via a statistical classifier; (3) DNN-based approaches (Ling et al., 2017; Wang et al., 2017), which map the given text into the corresponding math operation/equation via a DNN; and (4) Mixed approaches, which identify entities and quantities with rules, yet, decide operands and operations via statistical/DNN classifiers. This category can be further divided into two subtypes: (a) Without understanding (Roy and Roth, 2015; Koncel-Kedziorski et al., 2015; Huang et al., 2017; Shrivastava et al., 2017), which does not check the entity-attribute consistency between each quantity and the target of the given question; and (b) With understanding (Lin et al., 2015; Mitra and Baral, 2016; Roy and Roth, 2017), which also checks the entity-attribute consistency while solving the problem.

---
[1] It is a survey paper which reviews most of the rule-based approaches before 2008.

However, a widely covered rule-set is difficult to construct for the rule-based approach. Also, it is awkward in resolving ambiguity problem. In contrast, the performance of purely statistics-based approaches deteriorates significantly when the MWP includes either irrelevant information or information gaps (Hosseini et al., 2014), as it is solved without first understanding the meaning.

For the category (4a), since the physical meaning is only implicitly utilized and the result is not generated via inference, it would be difficult to explain how the answer is obtained in a human comprehensible way. Therefore, the categories (2), (3) and (4a) belong to the less favored direct translation approach[2] (Pape, 2004).

In contrast, the approaches of (4b) can avoid the problems mentioned above. However, among them, Mitra and Baral (2016) merely handled *Addition* and *Subtraction*. Only the *meaning-based framework* proposed by Lin et al. (2015) can handle general MWPs via understanding and reasoning. Therefore, it is possible to explain how the answer is obtained in a human comprehensible way (Huang et al., 2015). However, although their design looks promising, only a few Chinese MWPs had been tested and performance was not evaluated. Accordingly, it is hard to make a fair comparison between their approach and other state-of-the-art methods. In addition, in their prototype system, the desired operands of arithmetic operations are identified with predefined lexico-syntactic patterns and ad-hoc rules. Reusing the patterns/rules designed for Chinese in another language is thus difficult even if it is possible.

In this paper, we adopt the framework proposed by Lin et al. (2015) to solve English MWPs (for its potential in solving difficult/complex MWPs and providing more human comprehensible explanations). Additionally, we make the following improvements: (1) A new statistical model is proposed to select operands for arithmetic operations, and its model parameters can be automatically learnt via weakly supervised learning (Artzi and Zettlemoyer, 2013). (2) A new informative and robust feature-set is proposed to select the desired arithmetic operation. (3) We show the proposed approach significantly outperforms other existing systems on the common benchmark datasets reported in the literature. (4) A noisy dataset with more irrelevant quantities in MWPs is created and released. It could be used to check if an approach really understands what a given MWP looks for. (5) An experiment is conducted to compare various approaches on this new dataset. The superior performance of our system demonstrates that the proposed meaning-based approach has good potential in handling difficult/complex MWPs.

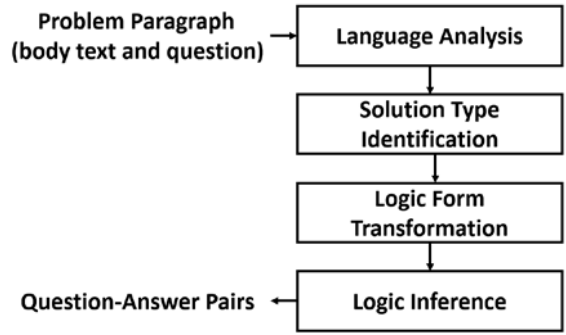

Figure 2: The diagram of *MeSys* framework

## 2 System Description

The adopted meaning-based framework (Lin et al., 2015) is a pipeline with following four stages (see Figure 2): (1) *Language Analysis*, (2) *Solution Type Identification*, (3) *Logic Form Transformation* and (4) *Logic Inference*. We use the Stanford CoreNLP suite (Manning et al., 2014) as the language analysis module. The other three modules are briefly described below. Last, we adopt the *weakly supervised learning* (Artzi and Zettlemoyer, 2013; Kushman et al., 2014) to automatically learn the model parameters without manually annotating each MWP with the adopted solution type and selected operands benchmark.

### 2.1 Solution Type Identification (STI)

After language analysis, each MWP is assigned with a specific solution type (such as *Addition*, *Multiplication*, etc.) which indicates the stereotype math operation pattern that should be adopted to solve this problem. We classify the English MWPs released by Hosseini et al. (2014) and Roy and Roth (2015) into 6 different types: *Addition*, *Subtraction*, *Multiplication*, *Division*, *Sum* and *TVQ-F*[3]. An SVM (Chang and Lin, 2011) is used to identify the solution type with 26 features. Most of them are derived from some important properties associated with each quantity.

---

[2] According to (Pape, 2004), the meaning-based approach of solving MWPs achieves the best performance among various behaviors adopted by middle school children.

[3] *TVQ-F* means to get the final state of a Time-Variant-Quantity that involves both Addition and Subtraction.

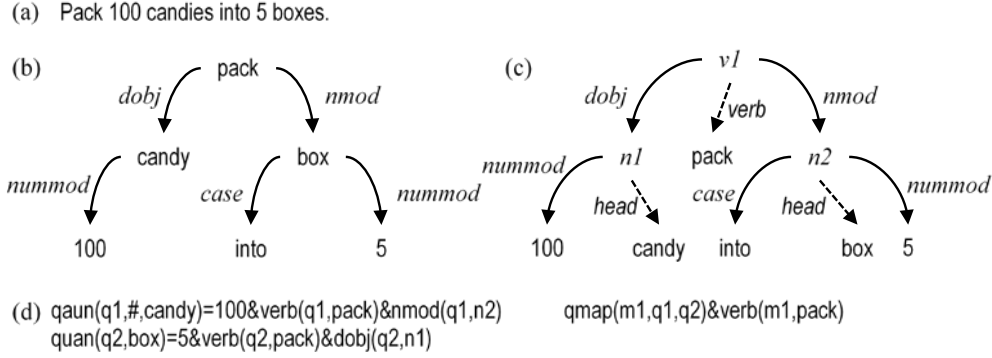

Figure 3: An example of logic form transformation

In addition to the properties *Entity*[4] and *Verb* (Hosseini et al., 2014) associated with the *quantity*, we also introduce a new property *Time* which encodes the tense and aspect of a verb into an integer to specify a point in the timeline. We assign 2, 4, and 6 to the tenses *Past*, *Present* and *Future*, respectively, and then adjust it with the aspect-values -1, 0 and 1 for *Perfect*, *Simple*, and *Progressive*, respectively.

Another property *Anchor* is associated with the *unknown quantity* asked in the question sentence. If the subject of the question sentence is a noun phrase (e.g., "*how many apples does **John** have?*"), *Anchor* is the *subject* (i.e., *John*). If the subject is an expletive nominal (e.g. "*how many apples are there in the box?*"), then *Anchor* is the associated nominal modifier *nmod* (i.e., "*box*"). Otherwise, *Anchor* is set to "*Unknown*".

Inspired by (Hosseini et al., 2014), we transform *Verb* to *Verb-Class* (*VC*) which is *positive*, *negative* or *stative*. A verb is *positive/negative* if it increases/decreases the associated quantity of the subject. For example, in the sentence "*Tom borrowed 3 dollars from Mike*", the verb is *positive* because the money of subject "*Tom*" increases.

However, a *positive* verb does not always imply the *Addition* operation. If the question is "*How much money does Mike have now?*" for the above body sentence, *the operation* should be *Subtraction*. Two new properties *Anchor-Role* (*AR*) and *Action* (*A*) are thus proposed: $AR_i$ indicates the role that *Anchor* associated with $q_i$, and is set to *nsubj/obj/nmod/$\phi$*. $A_i$ is determined by following rules: (1) $A_i$=*positive* if ($VC_i$, $AR_i$) is either (*positive*, *nsubj*) or (*negative*, *obj/nmod*). (2) $A_i$=*negative* if ($VC_i$, $AR_i$) is either (*negative*, *nsubj*) or (*positive*, *obj/nmod*). (3) Otherwise, $A_i$=$VC_i$.

To rule out the noisy quantities introduced by irrelevant information, we further associate each known quantity with the property *Relevance* (*R*) according to the unknown quantity asked in the question sentence. Let $q_i$ denote the *i*-th known quantity, $E_i$ denote the entity of $q_i$, $X_i$ denote the property *X* of $q_i$, $q_U$ denote the unknown quantity asked, and $X_U$ denote the property *X* of $q_U$. $R_i$ is specified with following rules: (1) $R_i$=2 (Directly-Related) if either {*Anchor* is *Unknown* & $E_i$ entails $E_U$} or {*Anchor* is not *Unknown* & $AR_i \neq \phi$ & $E_i$ entails $E_U$} (2) $R_i$=1 (Indirectly-Related) if there is a $q_j$ which maps[5] to $q_i$ and $R_j$=2 (i.e., $q_j$ is Directly-Related). (3) $R_i$=0 (Unrelated) otherwise.

The solution type is identified by an SVM based on 26 binary features. Let the symbols **p**, **n**, **s**, *A*, *E*, *R*, *T*, *V*, $S_B$, $S_Q$ and $w_Q$ stand for *positive*, *negative*, *stative*, *Action*, *Entity*, *Relevance*, *Time*, *Verb*, "*a body sentence*", "*the question sentence*" and "*a word in question sentence*" respectively. Also, let *I(x)* be the indicator function to check if *x* is true. The 26 features are briefly described as follows:

(1) $VC_U$=**p**; (2) $\exists R_i$=2 s.t. $A_i$=**p**; (3) $\exists R_i$=2 s.t. $A_i$=**n**;
(4) $\exists R_i$=2 s.t. $A_i$=**s**; (5) $\sum_i I(R_i=2) > 2$;
(6) $\sum_i I(R_i=2$ & $A_i \in \{$**p**, **n**$\}) = 2$;
(7) $\exists R_i$=2 s.t. $A_i$=**p** & $T_U < T_i$;
(8) $\exists R_i$=2 s.t. $A_i$=**n** & $T_U < T_i$;
(9) $\exists R_i$=2 s.t. $A_i$=**s** & $T_i$=max $T_j$;
(10) $\exists R_i$=2 s.t. $A_i$=**s** & $T_i < T_U$;
(11) $T_U \geq$ max $T_i$; (12) $T_U \leq$ min $T_i$;
(13) $\forall R_i$=2, $V_i$ are the same; (14) $\forall R_i$=2 s.t. $T_i$=$T_U$;
(15) $\forall R_i$=2, $T_i$ are the same;
(16) $\exists R_i$=2, $\exists R_j$=1 s.t. $q_i$ maps to $q_j$ & $q_i > q_j$;

---

[4] In our works, the term "*Entity*" also includes the unit of the quantity (e.g., "*cup of coffee*").

[5] That is, $q_i$ is linked to a directly-related quantity $q_j$ under an expression such as "*2 pencils weigh 30 grams*".

(a) A sandwich is priced at $0.75. A pudding is priced at $0.25. Tim bought 2 sandwiches and 4 puddings. Mary bought 2 puddings. How much money should Tim pay?

(b) …price(sandwich,0.75)&price(pudding,0.25)… quan(q1,#,sandwich)=2&verb(q1,buy)&nsubj(q1,Tim)… quan(q2,#,pudding)=4&verb(q2,buy)&nsubj(q2,Tim)… quan(q3,#,pudding)=2&verb(q3,buy)&nsubj(q3,Mary)… ASK Sum(quan(?q,dollar,#),verb(?q,pay)&nsubj(?q,Tim))

(c) quan(?q,?u,?o)&verb(?q,buy)&nsubj(?q,?a)&price(?o,?p) → quan($q,dollar,#)=quan(?q,?u,?o)×?p & verb($q,pay) & nsubj($q,?a)

(d) quan(q4,dollar,#)=1.5&verb(q4,pay)&nsubj(q4,Tim)… quan(q5,dollar,#)=1&verb(q5,pay)&nsubj(q5,Tim)… quan(q6,dollar,#)=0.5&verb(q6,pay)&nsubj(q6,Mary)

Figure 4: A logic inference example

(17) $\exists R_i=2, \exists R_j=1$ s.t. $q_i$ maps to $q_j$ & $q_i$ is associated with a word "each/every/per/a/an";
(18) $\exists R_i=2, \exists R_j=1$ s.t. $q_i$ maps to $q_j$ & $q_j$ is associated with a word "each/every/per/a/an";
(19) $\exists q_i, q_j, q_k$ s.t. $R_i = R_j = R_k = 2$ & $V_i = V_j = V_k$;
(20) $\exists w_Q \in \{total, in\ all, altogether, sum\}$;
(21) $\exists w_Q \in \{more, than\}$ or $\exists w_Q$ s.t. $w_Q$-POS=RBR;
(22) $\exists w_Q =$ "left"; (23). $\exists q_i$ appears in $S_Q$;
(24) "the rest V $E_U$" appears in $S_B$ (V for any verb);
(25) "each NN" appears in $S_Q$ (NN for any noun);
(26) $Anchor_U$ is $Unknown/nmod$ & $VC_U =$ s.

## 2.2 Logic Form Transformation (LFT)

The results of language analysis are transformed into a logic form, which is expressed with the *first-order logic* (FOL) formalism (Russell and Norvig, 2009). Figure 3 shows how to transform the sentence (a) "*Pack 100 candies into 5 boxes.*" into the corresponding logic form (d). First, the dependency tree (b) is transformed into the semantic representation tree (c) adopted by Lin et al., (2015). Afterwards, according to the procedure proposed in (Lin et al., 2015), the domain-dependent logic expressions are generated in (d).

The domain-dependent logic expressions are related to crucial generic math facts, such as *quantities* and *relations* between quantities. The FOL function $quan(quan_{id}, unit^6, entity)=number$ is for describing the quantity fact. The first argument denotes its unique *identifier*. The other arguments and the function value describe its meaning. Another FOL predicate $qmap(map_{id}, quan_{id1}, quan_{id2})$ (denotes the mapping from $quan_{id1}$ to $quan_{id2}$) is for describing a relation between two quantity facts, where the first argument is a unique identifier to represent this relation.

The role-tags (e.g., *verb*, *dobj*, etc.) associated with $quan_{id}$ and $map_{id}$ denote entity attributes (i.e., the physical meaning of the quantity), are created to help the logic inference module find the solution. For example, $quan(q_2,\#,box) = 5$ & $verb(q_2,pack)$ &… means that $q_2$ is the quantity of boxes being packed. With those role-tags, the system can select the operands more reliably, and the inference engine can also derive new quantities to solve complex MWPs which require multi-step arithmetic operations (see section 2.3).

The question in the MWP is also transformed into an FOL-like utility function according to the solution type to ask the logic inference module to find out the answer. For example, the utility function instance $Division(quan(q_1, \#, candy), quan(q_2, \#, box))$ asks the inference module to divide "*100 candies*" by "*5 boxes*". Since associated operands must be specified before calling those utility functions, a statistical model (see section 2.4) is used to identify the appropriate quantities.

## 2.3 Logic Inference

The logic inference module adopts the inference engine from (Lin et al., 2015). Figure 4 shows how it uses inference rules to derive new facts from the initial facts directly provided from the description. The MWP (a) provides some facts (b) generated from the LFT module. An inference rule (c)[7], which implements the common sense that people must pay money to buy something, is unified with the given facts (b) and derives new facts (d). The facts associated with $q_6$ can be interpreted as "*Mary paid 0.5 dollar for two puddings*".

The inference engine (IE) also provides 5 utility functions, including *Addition*, *Subtraction*, *Multiplication* and *Division*, and *Sum*. The first four utilities all return a value by performing the named math operation on its two input arguments. On the other hand, *Sum(function,condition)* returns the sum of the values of FOL *function instances* which can be unified with the first argument (i.e., *function*) and satisfy the second argument (i.e., *condition*). For example, according to

---

[6] This second argument denotes the associated unit used to count the entity. It is set to "#" if the unit of the entity is not specified.

[7] In the inference rule, $q is a meta symbol to ask the inference engine to generate a unique identifier for the newly derived quantity fact.

(a) Tim bought 2 roses and 3 lilies. Mary bought 4 roses and 5 lilies. How many flowers did Tim buy?

(b) quan(q1,#,rose)=2&verb(q1,buy)&nsubj(q1,Tim)…
quan(q2,#,lily)=3&verb(q2,buy)&nsubj(q2,Tim)…
quan(q3,#,rose)=4&verb(q3,buy)&nsubj(q3,Mary)…
quan(q4,#,lily)=5&verb(q4,buy)&nsubj(q4,Mary)…
quan(q$_Q$,#,flower)&verb(q$_Q$,buy)&nsubj(q$_Q$,Tim)…

Figure 5: An example for operand selection

the last line in Figure 4(b), three newly derived quantity facts *q4*, *q5* and *q6* (in 4(d)) can be unified with the first argument *quan(?q,dollar,#)* in 4(c), but only *q4* and *q5* satisfy the second argument *verb(?q,pay)&nsubj(?q,Tim)*. As a result, the answer 2.5 is returned by taking sum on the values of the quantity facts *quan(q4,dollar,#)* and *quan(q5,dollar,#)*.

## 2.4 Probabilistic Operand Selection

The most error-prone part in the LFT module is instantiating the utility function of math operation especially if many irrelevant quantity facts appear in the given MWP. Figure 5 shows the LFT module needs to select two quantity facts (among 4) for *Addition*. Please note that the *question quantity* $q_Q$, transformed from "*how many flowers*", is not a candidate for operand selection.

Lin et al., (2015) used predefined lexico-syntactic patterns and ad-hoc rules to instantiate utility functions. However, their rule-based approach fails when the MWP involves more quantities. Therefore, we propose a statistical model to select operands for the utility functions *Addition*, *Subtraction*, *Multiplication* and *Division*. The operand selection procedure can be regarded as finding the most likely configuration $(o_1^n, r)$, where $o_1^n = o_1, \cdots, o_n$ is a sequence of random indicators which denote if the corresponding quantity will be selected as an operand, and $r$ is a tri-state variable to represent the relation between the values of two operands (i.e., $r = -1, 0$ or $1$; which denote that the first operand is less than, equal to, or greater than the second operand, respectively). Given a solution type $s$, the MWP logic expressions $\mathbb{L}$ and the $n$ quantities $q_1^n = q_1, \cdots, q_n$ in $\mathbb{L}$. The procedure is formulated as:

$$P(r, o_1^n | q_1^n, \mathbb{L}, s) \approx P(r|s) \times P(o_1^n | q_1^n, \mathbb{L}, s), \quad (1)$$

$P(r|s)$ simply refers to *Relative Frequency* (as it has only a few parameters and we have enough training samples). $P(o_1^n | q_1^n, \mathbb{L}, s)$ is further derived as:

$$P(o_1^n | q_1^n, \mathbb{L}, s) \approx \prod_{i=1}^{n} P(o_i | q_i, \mathbb{L}, s) \approx \prod_{i=1}^{n} P(o_i | \Phi(q_i, \mathbb{L}, s)), \quad (2)$$

where $\Phi(\cdot)$ is a feature extraction function to map $q_i$ and its context into a feature vector. Here, the probabilistic factor $P(o_i | \Phi(q_i, \mathbb{L}, s))$ is obtained via an SVM classifier (Chang and Lin, 2011).

$\Phi(\cdot)$ extracts total 25 features (specified as follows, and 24 of them are binary) for $q_i$. The following 11 of them are independent on the question in the MWP:

1. Four features to indicate if $s$ is *Addition*, *Subtraction*, *Multiplication* or *Division*.
2. A feature to indicate if $q_i$ is within a *qmap*(…).
3. A feature to indicate if $q_i = 1$.
4. Five features to indicate if $n < 2, n = 2, n = 3, n = 4$ or $n > 4$; where $n$ is the number of quantities in Eq (1).

$\Phi(\cdot)$ also extracts features by matching the logic expressions of $q_i$ with those of *question quantity* $q_Q$ to check the role-tag consistencies between $q_i$ and $q_Q$. Another fourteen features are extracted with three indicator functions $I_m(\cdot)$, $I_e(\cdot)$, $I_\exists(\cdot)$ and one tri-state function $T_m(\cdot)$ as follows:

[ $I_m(q_i, q_Q, entity), I_e(q_i, q_Q, entity),$
$I_m(q_i, q_Q, verb), I_e(q_i, q_Q, verb),$
$I_\exists(q_Q, nsubj), T_m(q_i, q_Q, nsubj),$
$I_\exists(q_Q, modifier), I_m(q_i, q_Q, modifier),$
$I_\exists(q_Q, place), I_m(q_i, q_Q, place),$
$I_\exists(q_Q, temporal), I_m(q_i, q_Q, temporal),$
$I_\exists(q_Q, xcomp), I_m(q_i, q_Q, xcomp)$ ]

where the indicator functions $I_m(x, y, z)$ checks if the $z$ of $x$ matches the $z$ of $y$, $I_e(x, y, z)$ checks if the $z$ of $x$ entails the $z$ of $y$ and $I_\exists(y, z)$ checks if the $z$ of $y$ exists. $T_m(q_i, q_Q, nsubj)$ returns "exact-match" (if *nsubj* of $q_i$ matches *nsubj* of $q_Q$), "quasi-match" (if *nsubj* of $q_Q$ does not exist or is a plural pronoun), and "unmatch".

$I_e(\cdot)$ uses the WordNet hypernym and hyponym relationship to judge whether one entity/verb entails another one or not via checking if they are in an inherited hypernym-path in WordNet. The *entity*, *verb* and *nsubj* of a quantity are determined according to the logic expressions. The modifier, place, temporal and *xcomp* of a quantity are extracted from the dependency tree with some lexico-syntactic patterns. For example, the *modifier* and *place* of the quantity in the sentence "*There are 30 red flowers in the garden.*" are "*red*" and "*garden*" respectively. The *temporal* and *xcomp* of a quantity are extracted according to the dependency relations "*tmod*" (i.e., *temporal modifier*) and "*xcomp*" (i.e., *open clausal complement*), respectively.

## 3 Datasets for Performance Evaluation

The AI2 dataset provided by Hosseini et al. (2014) and the IL dataset released by Roy and Roth (2015) are adopted to compare our approach with other state-of-the-art methods. The AI2 dataset has 395 MWPs on addition and subtraction, with 121 MWPs containing irrelevant information (Hosseini et al., 2014). It is the most popular one for comparing different approaches. On the other hand, the IL dataset consists of 562 elementary MWPs which can be solved by one of the four arithmetic operations (i.e., +, −, ×, and ÷) without any irrelevant quantity. It is the first publicly available dataset for comparing performances that covers all four arithmetic operations.

However, the difficulty of solving an MWP depends not only on the number of arithmetic operations required, but also on how many irrelevant quantities inside, and even on how the quantities are described. One way to test if a proposed approach solves the MWPs with understanding is to check whether it is robust to those irrelevant quantities. Therefore, it is desirable to have a big enough dataset that contains irrelevant quantities which are created under different situations (e.g., confusing with an irrelevant agent, entity, or modifier, etc.) and allow us to probe the system weakness from different angles. We thus create a new dataset with more irrelevant quantities[8]. But before we do that, we need to know how difficult the task of solving the given MWPs is. Therefore, we first propose a way to measure how easy that a system solves the problem by simply guessing.

### 3.1 Perplexity-flavor Measure

We propose to adopt the *Perplexity* to measure the task difficulty, which evaluates how likely a solver will get the correct answer by guessing. Every MWP in the datasets can be associated with a solution expression template, such as "□ + □" or "□ − □", where the symbol □ represents a slot to hold a quantity. The solution can be obtained by placing correct quantities at appropriate slots. A random baseline is to solve an MWP by guessing. It first selects a solution expression template according to the prior distribution of the templates and then places quantities into the selected template according to the uniform distribution.

The expected accuracy of the random baseline on solving an MWP is a trivial combination and permutation exercise[9]. For example, the expected accuracy of solving an MWP associated with "□ + □" template is $p_{\Box+\Box} \times {_nC_2}^{-1}$, where the factor $p_{\Box+\Box}$ denotes the prior probability of the template "□ + □" and $n$ is the total number of quantities (including irrelevant ones) in the MWP. On the other hand, expected accuracy of solving an MWP associated with "□ − □"[10] template is $p_{\Box-\Box} \times {_nP_2}^{-1}$. Let $A_i$ denote the expected accuracy of solving the $i$-th MWP in a dataset. The accuracy of the random baseline on the dataset of size $N$ is then computed as $A = (1/N) \times \sum_{i=1}^{N} A_i$.

The word "*Accuracy*" comprises the opposite sense of the word "*Perplexity*"[11] (i.e., in the sense of how hard a prediction problem is). The lower the *Accuracy* is, the higher the *Perplexity* is. Therefore, we transform the *Accuracy* measure into a *Perplexity-Flavor* measure (PP) via the formula:

$$\text{PP} = 2^{-\log_2 A}$$

For instance, the *Perplexity-Flavor* measures of AI2 and IL datasets are 4.46 and 8.32 respectively.

### 3.2 Noisy Dataset

Human Math/Science tests have been considered more suitable for judging AI progress than Turing test (Clark and Etzioni, 2016). In our task, solving MWPs is mainly regarded as a test for intelligence (not just for creating a *Math Solver* package). By injecting various irrelevant quantities into original MWPs, a noisy dataset is thus created to assess if a solver solves the MWPs mainly via *understanding* or via *mechanical/statistical pattern matching*. If a system solves an MWP mainly via pattern matching, it would have difficulty in solving a similar MWP augmented from the original one with some irrelevant quantities. Therefore, we first create a noisy dataset by selecting some

---

[8] The IL dataset does not include any irrelevant information; on the other hand, the AI2 dataset only contains 121 MWPS with irrelevant information (but not systematically created).

[9] Let ${_nC_k}$ denote $k$-combinations of $n$ and ${_nP_k}$ denote $k$-permutations of $n$.

[10] We assume the operands have different values and, therefore, they are not permutable for the *subtraction* operator.

[11] The *Perplexity* of a uniform distribution over $k$ discrete events (such as casting a fair $k$-sided dice) is $k$.

(a) Tim has 10 yellow flowers and 12 red flowers. How many flowers does Tim have?
(a.1) Tim has … <u>Mary</u> has 3 yellow flowers. How many …
(a.2) Tim has … Tim also has 3 <u>books</u>. How many …

Figure 6: Examples of noisy sentences

MWPs that can be correctly solved, and then augmenting each of them with an additional noisy sentence which involves an irrelevant quantity. This dataset is created to examine if the solver knows that this newly added quantity is irrelevant.

Figure 6 shows how we inject noise into an MWP (a). (a.1) is created by associating an irrelevant quantity to a new subject (i.e., *Mary*). Here the ellipse symbol "…" denotes unchanged text. (a.2) is obtained by associating an irrelevant quantity to a new entity (i.e., *books*). In addition, we also change modifiers (such as *yellow*, *red*, …) to add new noisy sentence (not shown here). Since the noisy dataset is not designed to assess the lexicon coverage rate of a solver, we reuse the words in the original dataset as much as possible while adding new subjects, entities and modifiers.

136 MWPs that both Illinois Math Solver[12] (Roy and Roth, 2016) and our system can correctly solve are selected from the AI2 and IL datasets. This subset is denoted as OSS (Original Sub-Set). Afterwards, based on the 136 MWPs of OSS, we create a noisy dataset of 396 MWPs by adding irrelevant quantities. This noisy dataset is named as NDS[13]. Table 1 lists the size of MWPs, *Perplexities* (PP), and the average numbers of quantities in each MWP of these two datasets.

## 4 Experimental Results and Discussion

We compare our approach with (Roy and Roth, 2015) and (Roy and Roth, 2017) because they achieved the state-of-the-art performance on the IL dataset. In the approach of (Roy and Roth, 2015), each quantity in the MWP was associated with a *quantity schema* whose attributes are extracted from the context of the quantity. Based on the attributes, several statistical classifiers were used to select operands and determine the operator. They also reported the performances on the AI2 dataset to compare their approach with those of others (e.g., Kushman et al. (2014), which is a purely statistical approach that aligns the text with various pre-extracted equation templates). Roy and Roth (2017) further introduced the concept of *Unit Dependency Graphs* to reinforce the consistency of physical units among selected operands associated with the same operator.

|  | OSS | NDS |
|---|---|---|
| # MWPs | 136 | 396 |
| *Perplexity* (PP) | 7.42 | 18.83 |
| #Quantities | 2.64 | 3.62 |

Table 1: *Perplexity* measures of OSS and NDS

|  | AI2 | IL |
|---|---|---|
| Our system (Statistical) | **81.5** | **81.0** |
| Our system (DNN) | 69.8 | 70.6 |
| (Roy and Roth, 2017) | 76.2 | 74.4 |
| (Roy and Roth, 2015) | 78.0 | 73.9 |
| (Kushman et al., 2014) | 64.0 | 73.7 |

Table 2: Performances of various approaches

|  | AI2 | IL |
|---|---|---|
| STI (Statistical) | 83.0 | 83.1 |
| STI (DNN) | 74.5 | 68.8 |
| LFT | 92.1 | 94.8 |

Table 3: Performances of *different* STIs and LFT

To compare the performance of the statistical method with the DNN approach, we only implement a Bi-directional RNN-based *Solution Type Identifier* (as our original statistical *Operand Selector* is relatively much better). It consists of a word embedding layer (for both body and question parts), and a bidirectional GRU layer as an encoder. We apply the attention mechanism to scan all hidden state sequence of body by the last hidden state of question to pay more attention to those more important (i.e., more similar between the body and the question) words. Lastly, it outputs the solution type by a softmax function. We train it for 100 epochs, with mini-batch-size = 128 and learning-rate = 0.001; the number of nodes in the hidden layer is 200, and the drop-out rate is $0.7^{14}$.

We follow the same n-fold cross-validation evaluation setting adopted in (Roy and Roth, 2015) exactly. Therefore, various performances could be directly compared. Table 2 lists the accuracies of different systems in solving the MWPs of various datasets. The performance of (Roy and Roth, 2017) system is directly delivered by their code[15]. The last two rows are extracted from (Roy and Roth, 2015). The results show that our performances of the statistical approach significantly outperform that of our DNN approach and other systems on every dataset.

---

[12] We submit MWPs to Illinois Math Solver (https://cogcomp.cs.illinois.edu/page/demo_view/Math) in May and June, 2017.

[13] The noisy dataset can be downloaded from https://github.com /chaochun/nlu-mwp-noise-dataset. It includes 102 Addition, 147 Subtraction, 101 Multiplication and 46 Division MWPs.

[14] Since the dataset is not large enough for splitting a development set, we choose those hyper parameters based on the test set in coarse grain. Therefore, the DNN performance we show here might be a bit optimistic.

The performances of STI and LFT modules are listed in Table 3. As described in section 2, the benchmark for both solution type and the operand selection benchmark are automatically determined by weakly supervised learning. The first and second rows of Table 3 show the solution type accuracies of our statistical and DNN approaches, respectively. The third row shows the operand selection accuracy obtained by given the solution type benchmark. Basically, LFT accuracies are from 92% to 95%, and the system accuracies are dominated by STI. We analyzed errors resulted from our statistical STI on AI2 and IL datasets, respectively. For AI2, major errors come from: (1) failure of ruling out some irrelevant quantities (40%), and (2) making confusion between *TVQ-F* and *Sum* these two solution types (20%) for those cases that only involve addition operation (however, both types would return the same answer). For IL, major errors come from: (1) requiring additional information (35%), and (2) not knowing Part-Whole relation (17%). Table 4 shows a few examples for different STI error types.

The left-half of Table 5 shows the performances on the OSS and NDS datasets. Recall that OSS is created by selecting some MWPs which both Illinois Math Solver (Roy and Roth, 2016) and our system[16] can correctly solve. Therefore, both systems have 100% accuracy in solving the OSS dataset. However, these two systems behave very differently while solving the noisy dataset. The much higher accuracy of our system on the noisy dataset shows that our meaning-based approach understands the meaning of each quantity more. Therefore, it is less confused[17] with the irrelevant quantities.

One MWP in the noisy dataset that confuses *Illinois Math Solver* (IMS) is "*Tom has 9 yellow balloons. Sara has 8 yellow balloons. Bob has 5 yellow flowers. How many yellow balloons do*

---

[15] https://github.com/CogComp/arithmetic.
[16] In evaluating the performances on OSS and NDS datasets, our system is trained on the folds 2-5 of the IL dataset.
[17] Since the gap between two different types of approaches is quite big, those 396 examples on OSS and 196 examples on NDS are sufficient to confirm the conclusion.

| Error Type | Example |
|---|---|
| Confusing *TVQ-F* and *Sum* solution type | Sally found 9 seashells, Tom found 7 seashells, and Jessica found 5 seashells on the beach. How many seashells did they find together? |
| Requiring additional information | A garden has 52 rows and 15 columns of bean plans. How many plants are there in all? |
| Not knowing Part-Whole relationship | Eric wants to split a collection of peanuts into groups of 8. Eric has 64 peanuts. How many groups will be created? |

Table 4: Examples for different STI error types

|  | Statistical | R&R, 2016 |  | Statistical | DNN |
|---|---|---|---|---|---|
| OSS | 100 | 100 | OSS′ | 100 | 100 |
| NDS | 82.1 | 28.5 | NDS′ | 81.4 | 75.4 |

Table 5: Performances on the OSS and NDS

*they have in total?*", where the underlined sentence is the added noisy sentence. The solver sums all quantities and gives the wrong answer 22, which reveals that IMS cannot understand that the quantity "*5 yellow flowers*" is irrelevant to the question "*How many yellow balloons?*". On the contrary, our system avoids this mistake.

Although the meaning of each quantity is explicitly checked in our LFT module, our system still cannot correctly solve all MWPs in NDS. The error analysis reveals that the top-4 error sources are STI, LFT, CoreNLP and incorrect problem construction (for 27%, 27%, 18%, 18%), which indicates that our STI and LFT still cannot completely prevent the damage caused from the noisy sentences (which implies that more consistency check for quantity meaning should be done). The remaining errors are due to incorrect quantity extraction, lacking common-sense or not knowing entailment relationship between two entities.

A similar experiment is performed to check if the DNN approach will be affected by the noisy information more. We first select 124 MWPs (denoted as OSS′) from OSS that can be correctly solved by both our statistical and DNN approaches and then filter out 350 derived MWPs (denotes as NDS′) from NDS. The right-half of Table 5 shows that the performance of the DNN approach drops more than the statistical approach does in the noisy dataset, which indicates that our statistical approach is less sensitive to the irrelevant quantities and more close to human's approach.

## 5 Related Work

To the best of our knowledge, MWP solvers proposed before 2014 all adopted the rule-based approach. Mukherjee and Garain (2008) had given a good survey for all related approaches before 2008. Afterwards, Ma et al. (2010) proposed a MSWPAS system to simulate human arithmetic multi-step addition and subtraction behavior without evaluation. Besides, Liguda and Pfeiffer (2012) proposed a model based on augmented semantic networks, and claimed that it could solve multi-step MWPs and complex equation systems and was more robust to irrelevant information (also no evaluation).

Recently, Hosseini et al. (2014) proposed a *Container-Entity* based approach, which solved the MWP with a sequence of state transition. And Kushman et al. (2014) proposed the first statistical approach, which heuristically extracts some algebraic templates from labeled equations, and then aligns them with the given sentence. Since no semantic analysis is conducted, the performance is quite limited.

In more recent researches (Roy and Roth, 2015; Koncel-Kedziorski et al., 2015; Roy and Roth, 2017), quantities in an MWP were associated with attributes extracted from their contexts. Based on the attributes, several statistical classifiers were used to select operands and determine operators to solve multi-step MWPs. Since the physical meaning of each quantity is not explicitly considered in getting the answer, the reasoning process cannot be explained in a human comprehensible way. Besides, Shi et al. (2015) attacked the *number word problem*, which only deal with numbers, with a semantic parser. Mitra and Baral (2016) mapped MWPs into three types of problems, including Part-Whole, Change and Comparison. Each problem was associated with a generic formula. They used a log-linear model to determine how to instantiate the formula with quantities and solve the only one *Unknown* variable. They achieved the best performance on the AI2 dataset. However, their approach cannot handle *Multiplication* or *Division* related MWPs. Recently, DNN-based approaches (Ling et al, 2017; Wang et al, 2017) have emerged. However, they only attacked algebraic word problems, and required a very large training-set.

Our proposed approach mainly differs from those previous approaches in *combining the statistical framework with logic inference*, and also in *adopting the meaning-based statistical approach for selecting the desired operands*.

## 6 Conclusion

A *meaning-based* logic form represented with *role-tags* (e.g., *nsubj*, *verb*, etc.) is first proposed to associate the extracted math quantity with its physical meaning, which then can be used to identify the desired operands and filter out irrelevant quantities. Afterwards, a statistical framework is proposed to perform understanding and reasoning based on those logic expressions. We further compare the performance with a typical DNN approach, the results show the proposed approach is still better. We will try to integrate domain concepts into the DNN approach to improve the learning efficiency in the future.

The main contributions of our work are: (1) Adopting a meaning-based approach to solve English math word problems and showing its superiority over other state-of-the-art systems on common datasets. (2) Proposing a statistical model to select operands by explicitly checking the meanings of quantities against the meaning of the question sentence. (3) Designing a noisy dataset to test if a system solves the problems by understanding. (4) Proposing a perplexity-flavor measure to assess the complexity of a dataset.